\pdfoutput=1

\documentclass[11pt]{article}

\usepackage[]{ACL2023}

\usepackage{bbm}
\usepackage{todonotes}
\makeatletter
\newcommand*\iftodonotes{\if@todonotes@disabled\expandafter\@secondoftwo\else\expandafter\@firstoftwo\fi}  %
\makeatother

\definecolor{dandelion}{HTML}{FFD464}

\definecolor{bittersweet}{HTML}{C04F17}

\usepackage{tipa}
\newcommand{\ethz}{\text{\normalfont \textipa{D}}}
\newcommand{\mitinst}{\normalfont \text{\textipa{@}}}

\usepackage{times}
\usepackage{latexsym}
\usepackage{siunitx}
\usepackage{xspace}

\usepackage[T1]{fontenc}

\usepackage[utf8]{inputenc}

\usepackage{microtype}

\usepackage{inconsolata}

\usepackage[noend]{algpseudocode}
\usepackage{algorithm}
\usepackage{amsfonts}
\usepackage{amsmath}
\usepackage{amssymb}
\usepackage{bm}
\usepackage{cleveref}
\usepackage{amsthm}
\usepackage{mathtools}
\usepackage{relsize}
\usepackage{caption,subcaption,booktabs}
\usepackage{import}
\usepackage[htt]{hyphenat}

\usepackage{tikz}
\usetikzlibrary{positioning}
\usetikzlibrary{backgrounds}
\usetikzlibrary{calc}

\input{algos/algonames.tex}

\crefname{section}{\S}{\S\S}
\Crefname{section}{\S}{\S\S}
\crefname{table}{Tab.}{}
\crefname{figure}{Fig.}{}
\crefname{algorithm}{Algorithm}{}
\crefname{equation}{Eq.}{Eqs.}  %
\crefname{appendix}{App.}{}
\crefname{thm}{Theorem}{Theorems}
\crefname{prop}{Proposition}{Propositions}
\crefname{cor}{Corollary}{Corollaries}
\crefname{observation}{Observation}{Observations}
\crefname{assumption}{Assumption}{Assumptions}
\crefformat{section}{\S#2#1#3}

\theoremstyle{definition}
\newtheorem{definition}{Definition}[section]

\theoremstyle{definition}

\newcommand{\defn}[1]{{\textbf{#1}}}
\newcommand{\bert}{{BERT \xspace}}

\newcommand{\Daleth}{\text{$\daleth$}}

\newcommand{\rmd}{\mathrm{d}}
\newcommand{\defequals}{\triangleq}
\newcommand{\toppos}{\mathrm{top\_pos}}
\newcommand{\topneg}{\mathrm{top\_neg}}
\newcommand{\derivSemiring}[3]{\frac{\Daleth_{#1} #2}{\Daleth_{#1} #3}}
\newcommand{\deriv}[2]{\frac{\rmd{#1}}{\rmd{#2}}}
\newcommand{\derivMax}[2]{\derivSemiring{(\max, \times)}{#1}{#2}}
\newcommand{\derivEntropy}[2]{\derivSemiring{\text{Ent}}{#1}{#2}}
\newcommand{\derivGeneric}[2]{\derivSemiring{(\oplus,\otimes)}{#1}{#2}}

\usetikzlibrary{shapes.misc}
\newcommand{\thicctimes}{ \tikz[baseline=-.55ex] \node [inner sep=0pt,cross out,draw,line width=1pt,minimum size=1ex] (a) {};}
\newcommand{\parents}{\pi}

\DeclarePairedDelimiter{\ceil}{\lceil}{\rceil}

\newcommand{\K}{\mathbb{K}}
\newcommand{\R}{\mathbb{R}}

\def\calG{{\mathcal{G}}}

\def\calP{{\mathcal{P}}}

\def\calV{{\mathcal{V}}}

\def\ve{{\mathbf{e}}}

\def\vh{{\mathbf{h}}}

\def\vv{{\mathbf{v}}}

\def\vy{{\mathbf{y}}}

\newcommand{\grad}{\nabla}

\definecolor{red_fig}{HTML}{D95847}
\definecolor{blue_fig}{HTML}{5D7CE6}

\definecolor{skipcolor}{HTML}{CC78BC}
\definecolor{valuecolor}{HTML}{029E73}
\definecolor{querycolor}{HTML}{DE8F05}
\definecolor{keycolor}{HTML}{0173B2}

\newcommand{\mask}{\texttt{[MASK]}\xspace}

\title{Generalizing Backpropagation for Gradient-Based Interpretability}

\author{
Kevin Du$^{\ethz}$~\;~
Lucas Torroba Hennigen$^{\mitinst}$~\;~
Niklas Stoehr$^{\ethz}$~\;~\\
\textbf{Alexander Warstadt}$^{\ethz}$~\;~
\textbf{Ryan Cotterell}$^{\ethz}$
\\
$^{\ethz}$ETH Z{\"u}rich \quad $^{\mitinst}$MIT
\\
\normalsize 
\href{mailto:kevidu@ethz.ch}{\texttt{kevin.du@inf.ethz.ch}}~\;~ 
 \href{mailto:lucastor@mit.edu}{\texttt{lucastor@mit.edu}} 
~\;~\href{mailto:niklas.stoehr@inf.ethz.ch}{\texttt{niklas.stoehr@inf.ethz.ch}}\\
\normalsize 
\href{mailto:alexanderscott.warstadt@inf.ethz.ch}{\texttt{alexanderscott.warstadt@inf.ethz.ch}}~\;~ \href{mailto:ryan.cotterell@inf.ethz.ch}{\texttt{ryan.cotterell@inf.ethz.ch}}
}

\begin{document}
\maketitle
\begin{abstract}
Many popular feature-attribution methods for interpreting deep neural networks rely on computing the gradients of a model's output with respect to its inputs.
While these methods can indicate which input features may be important for the model's prediction, they reveal little about the inner workings of the model itself.
In this paper, we observe that the gradient computation of a model is a special case of a more general formulation using semirings. 
This observation allows us to generalize the backpropagation algorithm to efficiently compute other interpretable statistics about the gradient graph of a neural network, such as the highest-weighted path and entropy. 
We implement this generalized algorithm, evaluate it on synthetic datasets to better understand the statistics it computes, and apply it to study BERT's behavior on the subject--verb number agreement task (SVA). 
With this method, we (a) validate that the amount of gradient flow through a component of a model reflects its importance to a prediction and (b) for SVA, 
identify which pathways of the self-attention mechanism are most important.
\end{abstract}

\section{Introduction\footnote{Code and data available at \url{https://github.com/kdu4108/semiring-backprop-exps}.\looseness=-1}}
\label{sec:introduction}

One of the key contributors to the success of deep learning in NLP has been backpropagation~\citep{linnaianmaa_backprop}, a dynamic programming algorithm that efficiently computes the gradients of a scalar function with respect to its inputs~\citep{goodfellow2016deep}.
Backpropagation works by constructing a directed acyclic computation graph\footnote{With due care, a computation graph can be extended to the cyclic case.} that describes a function as a composition of various primitive operations, e.g., $+$, $\times$, and $\exp(\cdot)$, whose gradients are known, and subsequently traversing this graph in topological order to incrementally compute the gradients.
Since the runtime of backpropagation is linear in the number of edges of the computation graph, it is possible to quickly perform vast numbers of gradient descent steps in even the most gargantuan of neural networks.

While gradients are arguably most important for training, they can also be used to analyze and interpret neural network behavior. For example, feature attribution methods such as saliency maps~\citep{simonyan_2013_saliency_maps} and integrated gradients~\citep{sundararajan_2017_integrated_gradients} exploit gradients to identify which features of an input contribute most towards the model's prediction.
However, most of these methods provide little insight into how the gradient propagates through the computation graph, and those that do are computationally inefficient, e.g., \citet{lu-etal-inf-patterns-2021} give an algorithm for computing the highest-weighted gradient path that
runs in exponential time.\looseness=-1

In this paper, we explore whether examining various quantities computed from the gradient graph of a network, i.e., the weighted graph whose edge weights correspond to the local gradient between two nodes, can lead to more insightful and granular analyses of network behavior than the gradient itself.
To do so, we note that backpropagation is an instance of a shortest-path problem~\citep{mohri_semirings} over the $(+, \times)$ semiring.
This insight allows us to generalize backpropagation to other semirings, allowing us to compute statistics about the gradient graph beyond just the gradient, all while retaining backpropagation's linear time complexity.\footnote{
This is analogous to how, in the context of probabilistic context-free grammars, the inside algorithm can be modified to obtain the CKY algorithm~\citep{collins-inside-cky},
and, in the context of graphical models, how the sum-product algorithm for partition functions can be generalized to the max-product algorithm for MAP inference~\citep{graphical-models}.}

In our experiments, the first semiring we consider is the max-product semiring, which allows us to identify paths in the computation graph which carry most of the gradient, akin to \citeposs{lu-etal-inf-patterns-2021} influence paths.
The second is the entropy semiring \cite{eisner-2002-parameter},\footnote{\citet{eisner-2002-parameter} refers to this as the expectation semiring.} which summarizes how dispersed the gradient graph is, i.e., whether the gradient flows in a relatively focalized manner through a small proportion of possible paths or in a widely distributed manner across most paths in the network.
With experiments on synthetic data, we validate that the max-product semiring results in higher values for model components we expect to be more critical to the model’s predictions, based on the design of the Transformer~\citep{vaswani-transformer} architecture.
We further apply our framework to analyze the behavior of \bert\citep{bert} in a subject--verb agreement task \cite[SVA;][]{linzen_assessing_2016}.
In these experiments, we find that the keys matrix for subject tokens carries most of the gradient through the last layer of the self-attention mechanism.
Our results suggest that semiring-lifted gradient graphs can be a versatile tool in the interpretability researcher's toolbox.

\section{Gradient-based interpretability} 
\label{sec:related-work}
Neural networks are often viewed as black boxes because 
their inner workings are too complicated for a user to understand \textit{why} the model produced a particular prediction for a given input.
This shortcoming has spawned an active field of research in developing methods to better understand and explain how neural networks work.
For example, feature attribution methods
aim to measure the sensitivity of a model's predictions to the values of individual input features.
Many of these methods quantify feature attribution as the gradient of the model's output with respect to an input feature~\citep{simonyan_2013_saliency_maps, smilkov_2017_smoothgrad, sundararajan_2017_integrated_gradients}.
We note that while the general reliability and faithfulness of gradient-based methods has been a contentious area of research
\citep{adebayo_2018_sanity_checks_saliency, yona_2021_revisiting_sanity_checks, amorim_2023_evaluating_saliency}, gradient-based methods have nonetheless continued to be widely used \citep{han_2020_influence_functions_bert, supekar_2022_ig_application, novakovsky_2022_obtaining}.\looseness=-1

Other works have applied feature attribution methods to not only highlight sensitive input features but also uncover important internal neurons.
\citet{leino_2018_influence} define {influence} as the gradient of a quantity of interest with respect to a neuron, averaged across a collection of inputs of interest.
\citet{lu-etal-2020-influence} further
define and analyze the notion of {influence paths}, i.e., paths in the computation graph between the neuron of interest and the output that on average carry most of the gradient.
By applying this method to analyze the behavior of \citeposs{gulordava-etal-2018-colorless} LSTM language model on the SVA task, they draw conclusions about which internal components of the LSTM are most sensitive to the concept of number agreement based on the paths with the greatest amount of influence.

However, \citeposs{lu-etal-2020-influence} method
exhaustively enumerates all paths in the computation graph and ranks them by the amount of influence along each one. As the number of paths in a computation graph is usually exponential in the depth of a neural network, this quickly becomes intractable for larger networks \citep{lu-etal-inf-patterns-2021}. Therefore, this method is limited to computing influence paths for networks with very small numbers of paths. Indeed, while \citet{lu-etal-2020-influence} computed the influence along \num{40000} paths for a 2-layer LSTM, follow-up work that attempted to apply this method to \bert had to use an approximation which might not find the correct paths~\citep{lu-etal-inf-patterns-2021}.
The method we propose does not exhibit this issue and scales to any network one can train using backpropagation.

\section{Generalizing backpropagation}
\label{sec:generalizing-backpropagation}

\tikzset{%
    inputnode/.style={circle, draw=black, fill=blue!20},
    intermediatenode/.style={circle, draw=black},
    outputnode/.style={circle, draw=black, fill=red!20},
    background rectangle/.style={draw=white!15},
    show background rectangle,
}

\begin{figure*}
    \begin{subfigure}[b]{0.2\textwidth}
    \raisebox{0.9\height}{ 
        \begin{minipage}{\linewidth}
        \begin{align*}
             x_0 &= x \\
             x_1 &= y \\
             x_2 &= \exp(x_0) \\
             x_3 &= x_0 - x_1 \\
             x_4 &= x_1 \times x_3 \\
             x_5 &= x_2 + x_4 \\
        \end{align*}
        \end{minipage}%
        }
    \end{subfigure}
    \hfill
    \begin{subfigure}[b]{0.3\textwidth}
        \begin{tikzpicture}
        
        \node[inputnode] (x) {$x_0$};
        \node[inputnode] (y) [below=1cm of x] {$x_1$};
        
        \node[intermediatenode] (ex) [right=of x] {$x_2$};
        \node[intermediatenode] (xplusy) [right=of y] {$x_3$};
        
        \node[intermediatenode] (timesy) [right=of xplusy] {$x_4$};
        
        \node[outputnode] (f) [above=of timesy] {$x_5$};
    
        \draw[->] (x) -- (ex);
        \draw[->] (x) -- (xplusy);
        \draw[->] (y) -- (xplusy);
        \draw[->] (xplusy) -- (timesy);
        \path[->] (y) edge [bend right=60] node {} (timesy);
        \draw[->] (ex) -- (f);
        \draw[->] (timesy) -- (f);
        
        \end{tikzpicture}
        \caption{Computation graph of $f(x, y)$}
        \label{fig:computation-graph-example}
    \end{subfigure}
    \hfill
    \begin{subfigure}[b]{0.3\textwidth}
        \begin{tikzpicture}
        
        \node[inputnode] (x) {$x_0$};
        \node[inputnode] (y) [below=1cm of x] {$x_1$};
        
        \node[intermediatenode] (ex) [right=of x] {$x_2$};
        \node[intermediatenode] (xplusy) [right=of y] {$x_3$};
        
        \node[intermediatenode] (timesy) [right=of xplusy] {$x_4$};
        
        \node[outputnode] (f) [above=of timesy] {$x_5$};
    
        \draw[->] (x) to node [above, sloped] {$e^x$} (ex);
        \draw[->] (x) to node [above, sloped] {$1$} (xplusy);
        \draw[->] (y) to node [above, sloped] {$-1$} (xplusy);
        \draw[->] (xplusy) to node [above, sloped] {$y$} (timesy);
        \path[->] (y) edge [bend right=60] node [above] {$x - y$} (timesy);
        \draw[->] (ex) to node [above, sloped] {$1$} (f);
        \draw[->] (timesy) to node [above, sloped] {$1$} (f);
        
        \end{tikzpicture}
        \caption{Gradient graph of $f(x,y)$}
        \label{fig:gradient-graph-example}
    \end{subfigure}
    \caption{Comparison of the computation graph~(\cref{fig:computation-graph-example}) and gradient graph~(\cref{fig:gradient-graph-example}) of $f(x, y) = e^x + (x - y) y$. The expressions and primitives associated with each node in the graphs are shown on the left. Notice that the gradient graph is a labeled variant of the computation graph. Input nodes are shown in blue, and output nodes in red.}
    \label{fig:computation-vs-gradient-graph}
    \vspace{-0.5cm}
\end{figure*}
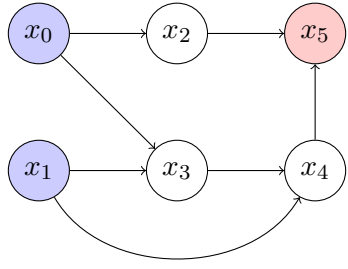
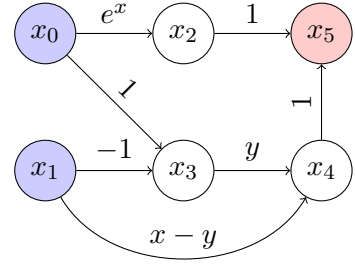

In this section, we build toward our generalization of backpropagation as a semiring-weighted dynamic program. 
At a high level, we observe that if we replace the addition and multiplication operations in the typical backpropagation algorithm with similar operations that satisfy the necessary properties, then the resulting algorithm will compute other useful statistics about the network's gradient graph in the same runtime as backpropagation.
In the remainder of this section, we make this notion of swapping operations precise by formulating backpropagation as a semiring algorithm, and later in \cref{sec:gradient_interpretability} we describe how different semirings yield different, useful, views of the gradient graph.

\subsection{Computation graphs}\label{sec:computation_graph}
Many classes of functions, e.g., machine learning models, can be expressed as compositions of differentiable functions.
Such functions can described by a computation graph~\citep{goodfellow2016deep}.
A \defn{computation graph} is an ordered\footnote{We require that nodes be ordered since primitives might not be invariant to permutations of their arguments, e.g., $\frac{x}{y} \neq \frac{y}{x}$ in general.} directed acyclic graph (DAG) where every node is associated with the application of a primitive operation, e.g., $+$, $\times$, and $\exp(\cdot)$, to the parents of that node.
These primitives all share the property that their gradients have a closed form and are assumed to be computable in constant time for the sake of analysis.
Source nodes in the graph are called \defn{input nodes}, and every computation graph has a designated \defn{output node} that encapsulates the result of the function.\footnote{For simplicity, we only consider scalar-valued functions, but extensions to vector-valued functions are possible and indeed commonplace in the literature.}
An example computation graph is shown in \cref{fig:computation-graph-example}.

If all input nodes are assigned a value, then one can perform a \defn{forward pass}, which calculates the value of the function at those inputs by traversing the graph in a topological order,\footnote{A topological ordering of a DAG is an ordering of its nodes such that node $i$ precedes node $j$ iff $i$ is not a child of $j$.} evaluating the values of each node until we reach the output node.
This procedure is shown in \cref{alg:forward_pass}.

\begin{algorithm}
\caption{  
    Forward-propagation
}\label{alg:forward_pass}
\begin{algorithmic}[1]
\Func{\forward($G$, $D$, $\tau$)}
\LineComment{$G$ is a computation graph with topologically-sorted nodes $V=[v_1, \ldots, v_N]$.}
\LineComment{$D$ is an ordered dictionary mapping from nodes to their values, with $D[v_i]$ initialized to the input value associated with $v_i \forall i \in [1, \ldots, m]$.}
\LineComment{$\tau : (V, V) \rightarrow \mathbb{N}$ is a function that maps a parent node to the index of the argument list of a function corresponding to a node. That is, given a node $v$ and parent node $u$, $\tau$ maps to an index in $\{1, \ldots, |\parents(v)|\}$ for all $v \in V, u \in \parents(v)$.}
\For{$k = m + 1, \ldots, N$} 
    \State $(\boldsymbol{a}_k)_{\tau(v_k, u)} \gets (D[u])_{u \in \parents(v_k)}$ \InlineComment{Retrieve the value for each input $u$ and store in the ordered argument tuple $\boldsymbol{a}_k$}

    \State $D[v_k] \gets f_k(\boldsymbol{a}_k)$
\EndFor
\State \Return $D$
\EndFunc
\end{algorithmic}
\end{algorithm}

\subsection{Backpropagation}\label{sec:backprop}

Encoding a function as a computation graph is useful because it enables the efficient computation of its gradients via automatic differentiation~\citep{griewank}.
Let $G$ be a computation graph with topologically sorted nodes $v_1, \ldots, v_N$, where $v_N$ is its output node.
The goal of \defn{automatic differentiation} is to compute $\deriv{v_N}{v_i}$ for some node $v_i$ in $G$.
\citet{bauer_paths} shows that $\deriv{v_N}{v_i}$ can be expressed as: 
\begin{equation}
    \deriv{v_N}{v_i} = \sum_{p \in \calP(i, N)}\prod_{(j, k) \in p} \deriv{v_k}{v_j}
    \label{eq:bauer-equation}
\end{equation}
where $\calP(i, N)$ denotes the set of \defn{Bauer paths}---directed paths in the computation graph $G$ from node $v_i$ to node $v_N$.\footnote{A directed path is an ordered set of node pairs, i.e., $\langle (i_1, i_2), (i_2, i_3), \ldots, (i_{p - 1}, i_p)\rangle$ where the second element of each pair matches the first element of the subsequent pair.}
That is, the gradient of the output $v_N$ with respect to a node $v_i$ equals the sum of the gradient computed along every path between $v_i$ and $v_N$, where the gradient along a path is the product of the gradient assigned to each edge along that path.
The gradient of each edge is easy to compute, as it corresponds to the gradient of a primitive.
To distinguish the original, unweighted computation graph from its gradient-weighted counterpart, we call the latter the \defn{gradient graph} $\calG(\cdot)$ of a function;
an example is shown in \cref{fig:gradient-graph-example}.
Note that this is a function of the input nodes, since the edge gradients are dependent on the input nodes.

In general, na{\"i}vely computing \cref{eq:bauer-equation} term by term is intractable since $\calP(i, N)$ can be exponential in the number of nodes in the computation graph.
By leveraging the distributivity of multiplication over addition, \defn{backpropagation}\footnote{Also known as reverse-mode automatic differentiation.} uses dynamic programming and the caching of intermediate values from the forward pass to compute \cref{eq:bauer-equation} in $O(|E|)$ time, where $|E|$ is the number of edges in $G$~\citep[p.~206]{goodfellow2016deep}. 
Backpropagation can be seen as traversing the computation graph in reverse topological order and computing the gradient of the output node with respect to each intermediate node until $v_i$ is reached.\footnote{Another efficient algorithm for computing \cref{eq:bauer-equation} is forward-mode automatic differentiation, which is most useful when one has more output nodes than input nodes in the network~\citep{griewank}. Since our formulation assumes a single output node, we focus solely on backpropagation.}

\subsection{Semiring backpropagation}
\label{sec:semiring-backprop}

The crucial observation at the core of this paper is that backpropagation need not limit itself to addition and multiplication: If, instead, we replace those operations with other binary operators that also exhibit distributivity, say $\oplus$ and $\otimes$, then this new algorithm would compute:
\begin{equation}
    \derivGeneric{v_N}{v_i} \defequals
    \bigoplus_{p \in \calP(i, N)} \bigotimes_{(j, k) \in p} \deriv{v_k}{v_j}
    \label{eq:semiring-bauer-equation}
\end{equation}
Clearly, the interpretation of this resulting quantity depends on how $\oplus$ and $\otimes$ are defined.
We discuss different options in \cref{sec:gradient_interpretability}, and in the remainder of this section we focus on how $\oplus$ and $\otimes$ have to behave to make them suitable candidates for replacement.

To make this notion more rigorous, we first need to introduce the notion of a semiring.
\begin{definition}
A \defn{semiring} (over a set $\K$) is an algebraic structure $(\K, \oplus, \otimes, \bar0, \bar1)$ such that:
\begin{enumerate}
\itemsep-0.3em 
\item $\oplus$: $\K{\times}\K\rightarrow\K$ is a commutative and associative operation  with identity element $\bar{0}$;
\item $\otimes$: $\K{\times}\K\rightarrow\K$ is an associative operation with identity element $\bar{1}$;
\item $\otimes$ distributes over $\oplus$;
\item $\bar0$ is an annihilator, i.e., for any $k \in \K$, $k \otimes \bar0 = \bar0 = \bar0 \otimes k$.
\end{enumerate}
\end{definition}

If we replace the operations and identity elements in backpropagation according to the semiring identities and operations, we obtain \defn{semiring backpropagation}, shown in \cref{alg:backprop_semirings}.
Regular backprogation amounts to a special case of the algorithm when run on the \defn{sum-product semiring} $(\R, +, \times, 0, 1)$.

\begin{algorithm}[h]
\caption{  
    Semiring backpropagation \\
    This algorithm is executed after the forward pass of a computation graph.
}\label{alg:backprop_semirings}
\begin{algorithmic}[1]

\Func{\backprop($G$, $D$)}
\LineComment{$G$ is a computation graph with topologically-sorted nodes $V=[v_1, ..., v_N]$.}
\LineComment{$D$ is an ordered dictionary mapping from node $v_i$ to its value, $\forall i \in [1, ..., m]$, computed by the forward pass}
\For{$v \in V$} %
    \State $B[v] \gets \bar{0}$ \InlineComment{$B$ is a dictionary mapping from $v_i$ to $\derivGeneric{v_N}{v_i}$}
\EndFor
\State $B[v_N] \gets \bar{1}$
\For{$i = N, ..., 1$} %
    \For{$u$ in $\parents(v_i)$} 
        \State $B[u] \gets B[u] \oplus \left(\deriv{v_i}{u} \Bigr\rvert_{D[u]} \otimes B[v_i] \right)$
    \EndFor
\EndFor
\State \Return $B$
\EndFunc
\end{algorithmic}
\hrulefill \\ 
For standard backpropagation, let $\oplus$ be the addition ($+$) operator and $\otimes$ be the times ($\times$) operator.
\end{algorithm}

\paragraph{Aggregated derivative.} 
\Cref{eq:semiring-bauer-equation} defines $\derivGeneric{v_N}{v_i}$ 
for a single node $v_i$.
However, often it is useful to aggregate this quantity across a set of nodes.
For example, when a token is embedded into a $d$-dimensional vector, each of its dimensions corresponds to a node in the computation graph, say $\calV = \{v_1, \ldots, v_{d}\}$.
Then, $\derivGeneric{v_N}{v_{j}}$ for the $j^\text{th}$ component of the representation does not capture the semiring-derivative with respect to the \emph{entire} representation of the token.
Hence, we define the \defn{aggregated derivative} with respect to a set of nodes $\mathcal{V}$ as:\footnote{This is equivalent to adding a dummy source node $v_0$ with outgoing edges of weight $\bar{1}$ to each node $v \in \mathcal{V}$ to the gradient graph and computing $\derivGeneric{v_N}{v_0}$.}\looseness=-1

\begin{equation}
    \derivGeneric{v_N}{\mathcal{V}} \defequals
    \bigoplus_{v \in \mathcal{V}} \derivGeneric{v_N}{v}
    \label{eq:total-semiring-derivative}
\end{equation}

\section{Interpreting semiring gradients}
\label{sec:gradient_interpretability}

In \cref{sec:generalizing-backpropagation}, we showed how to generalize backpropagation to the semiring case.
For any semiring of our choosing, this modified algorithm will compute a different statistic associated with a function's gradient.
We begin by motivating the standard $(+, \times)$ semiring which is common in the interpretability literature, before discussing the implementation and interpretation of the max-product and entropy semirings we focus on in this work.

\subsection{What is a $(+, \times)$ gradient?}\label{sec:og-gradient}
We start by reviewing the gradient interpretation in the $(+, \times)$ semiring, which corresponds to the standard definition of the gradient.
We explain why and how the gradient can be useful for interpretability.
Let $f : \R^D \rightarrow \R$ be a function differentiable at $\vy \in \R^D$ (e.g., a neural network model).
The derivative of $f$ at $\vy$, $\grad f(\vy)$, can be interpreted as the best linear approximation of the function at $\vy$ \citep{rudin1976principles}, viz., for any unit vector $\vv \in \R^D$ and scalar $\epsilon > 0$, we have: 
\begin{align}
    f(\vy + \epsilon \vv) = f(\vy) + \grad f(\vy)^\top (\epsilon \vv) + o(\epsilon)
\end{align}
As such, one can view gradients as answering \emph{counterfactual} questions:
If we moved our input $\vy$ in the direction $\vv$ for some small distance $\epsilon$, what is our best guess (relying only on a local, linear approximation of the function) about how the output of the model would change?\footnote{Indeed, this locality is a common source of criticism for gradient-based interpretability metrics as discussed in \cref{sec:related-work}.}

Gradient-based methods (as discussed in \Cref{sec:related-work}) are useful to interpretability precisely because of this counterfactual interpretation.
In using gradients for interpretability, researchers typically implicitly consider $\vv = \ve_i$, i.e., the $i^{\text{th}}$ natural basis vector, which approximates the output if we increment the model's $i^\text{th}$ input feature by one.
We can then interpret the coordinates of the gradient as follows: If its $i^{\text{th}}$ coordinate is close to zero, then we can be reasonably confident that small changes to that specific coordinate of the input should have little influence on the value of $f$.
However, if the gradient's $i^{\text{th}}$ coordinate is large in magnitude (whether positive or negative), then we may conclude that small changes in the $i^{\text{th}}$ coordinate of the input \emph{should} have a large influence on the value of $f$.

The subsequent two sections address a shortcoming in exclusively inspecting the gradient, which is fundamentally an aggregate quantity that sums over all individual Bauer paths.
This means, however, that any information about the structure of that path is left out, e.g., whether a few paths' contributions dominate the others.
The semiring gradients that we introduce in the sequel offer different angles of interpretation of such counterfactual statements.

\subsection{What is a $(\max, \times)$ gradient?}

While the $(+, \times)$ gradient has a natural interpretation given by calculus and has been used in many prior works~\citep{simonyan_2013_saliency_maps, bach_2015_lrp, sundararajan_2017_integrated_gradients} to identify input features that are most sensitive to a model's output, it cannot tell us \emph{how} the gradient flows through a gradient graph, as discussed in \Cref{sec:og-gradient}.
One way to compute a different quantity is to change the semiring.
The \defn{max-product semiring} $(\R \cup \{-\infty, \infty\}, \max, \times, -\infty, 1)$ is an enticing candidate:
In contrast to the $(+, \times)$ semiring, computing the gradient with respect to the $(\max, \times)$ semiring can help illuminate \emph{which} components of the network are most sensitive or critical to the model's input.
The $(\max, \times)$ gradient specifically computes the gradient along the Bauer path that has the highest value.
We term this path the \defn{top gradient path} in the sequel.
Formally, the $(\max, \times)$ gradient between $v_i$ and $v_N$
is:
\begin{equation}
    \derivMax{v_N}{v_i} \defequals
    \max_{p \in \mathcal{P}(i, N)}\prod_{(j, k) \in p}{\deriv{v_k}{v_j}}
\end{equation}
Note that variants of this definition are possible, e.g., we could have considered the \emph{absolute} values of the gradients $\left|\deriv{v_k}{v_j}\right|$ if we did not care about the overall impact as opposed to the most \emph{positive} impact on the output $v_N$.

The top gradient path can be used to examine branching points in a model's computation graph.
For example, 
in Transformer~\citep{vaswani-transformer} models, the input to an attention layer branches when it passes through both the self-attention mechanism and a skip connection. 
The input further branches within the self-attention mechanism between the keys, values, and queries (see \cref{fig:bert_kvq} for an illustration).
By examining the top gradient path at this branching point, we can identify not only whether the skip connection or self-attention mechanism is more critical to determining input sensitivity, but also which component within the self-attention mechanism itself (keys, queries, or values) carries the most importance.

\paragraph{Implementation.}
By using the max-product semiring in the backpropagation algorithm, we can compute the top gradient path in $O(|E|)$ time, where $|E|$ is the number of edges in the computation graph \cite[p.~206]{goodfellow2016deep}. 
See \cref{app:top_grad_path} for more details.

\subsection{What is an entropy gradient?}
\label{sec:entropy}

In addition to identifying the single top gradient path, it is also helpful to have a more holistic view of the gradient paths in a graph.
In particular, we may be interested in the path entropy of the gradient graph, i.e., the dispersion of the magnitudes of the path weights. 
Formally, for an input $\vy$ and its corresponding gradient graph $\calG(\vy)$ with nodes $v_1, \ldots, v_N$, the \defn{entropy} of all paths between $v_i$ and $v_N$ is defined as:
\begin{equation}
    \derivEntropy{v_N}{v_i} \defequals -\hspace{-0.3cm}\sum_{p \in \mathcal{P}(i, N)} \left|\frac{\pathgrad(p)}{Z}\right| \log \left|\frac{\pathgrad(p)}{Z}\right|
\end{equation}
where $\pathgrad(p) \defequals \prod_{(j, k) \in p}{\deriv{v_k}{v_j}}$ is the gradient of path $p$ and $Z = \sum_{p \in \mathcal{P}(i, N)} \left|\pathgrad(p)\right|$ is a normalizing factor.

Intuitively, under this view, the gradient graph $\calG(\cdot)$ encodes an (unnormalized) probability distribution over paths between $v_i$ and $v_N$ where the probability of a given path is proportional to the absolute value of the product of the gradients along each edge.
The entropy then describes the dispersion of the gradient's flow through all the possible paths in the graph from $v_i$ to $v_N$. 
For a given graph, the entropy is greatest when the gradient flows uniformly through all possible paths, and least when it flows through a single path.

\paragraph{Implementation.}
\citet{eisner-2002-parameter} proposed to efficiently compute the entropy of a graph by lifting the graph's edge weights into the \defn{expectation semiring} $(\R \times \R, \oplus, \otimes, \bar0, \bar1)$ where $\bar0 = \langle 0, 0 \rangle$, $\bar1 = \langle 1, 0 \rangle$ and:
\begin{itemize}
\itemsep-0.3em
    \item $\oplus$: $\langle a, b \rangle \oplus \langle c, d \rangle = \langle a + c, b + d \rangle$
    \item $\otimes$: $\langle a, b \rangle \otimes \langle c, d \rangle = \langle ac, ad + bc \rangle$
\end{itemize}

To leverage the expectation semiring, we first lift the weight of each edge in the gradient graph from $w$ to $\langle \left|w\right|, \left|w\right|\log \left|w\right|\rangle$ 
(where $w$ is the local derivative between two connected nodes in the gradient graph).
Then, by computing:
\begin{align}
    &\langle Z, -\hspace{-0.3cm} \sum_{p \in \mathcal{P}(i, N)} \hspace{-0.2cm} \left|\pathgrad(p)\right|\log\left|\pathgrad(p)\right| \rangle \\
    &= \bigoplus_{p \in \mathcal{P}(i, N)}\bigotimes_{(j, k) \in p}{\left\langle \left|\deriv{v_k}{v_j}\right|, -\left|\deriv{v_k}{v_j}\right|\log \left|\deriv{v_k}{v_j}\right|\right\rangle}\nonumber
\end{align}
in linear time using \Cref{alg:backprop_semirings}, we obtain $\langle Z, \sum_{p \in \mathcal{P}(i, N)} \left|\pathgrad(p)\right|\log\left|\pathgrad(p)\right| \rangle$, which are the normalizing factor and the unnormalized entropy of the graph, respectively.
As shown by \citet{li-eisner-2009-first}, we can then compute $\derivEntropy{v_N}{v_i} = \log Z - \frac{1}{Z}\sum_{p \in \mathcal{P}(i, N)} \left|\pathgrad(p)\right|\log\left|\pathgrad(p)\right|$.

\section{Experiments}
\label{sec:experiments}
To demonstrate the utility of semiring backpropogation, we empirically analyze their behavior on two simple transformer models (1-2 layers) on well-controlled, synthetic tasks.
We also explore semiring backpropogation on a larger model, \bert\citep{bert}, on the popular analysis task of subject--verb agreement to understand how our method can be useful for interpreting language models in more typical settings.

To implement semiring backpropagation, we developed our own Python-based reverse-mode automatic differentiation library, building off of the pedagogical library Brunoflow~\citep{Brunoflow} and translating it into JAX~\citep{jax2018github}.\footnote{Library available at \url{https://github.com/kdu4108/brunoflow}.}

\subsection{Validation on a synthetic task}
\label{sec:top_path_synthetic_experiments}

\begin{figure}[t]
     \centering
     \includegraphics[width=1.0\linewidth]{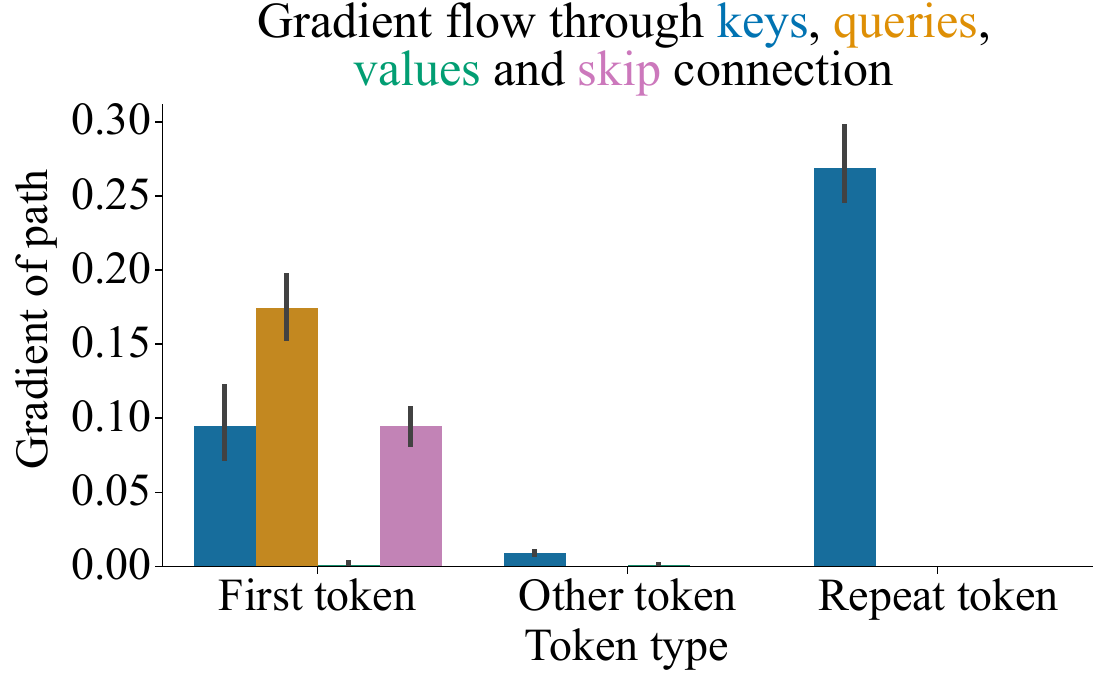} 
     \caption{High gradient flow through the \textcolor{querycolor}{queries} for the first token and \textcolor{keycolor}{keys} of the repeat token match the expected important components of self-attention for each token type, respectively.
     }
     \label{fig:max_grad_synthetic}
     \vspace{-0.5cm}
 \end{figure}
 
\paragraph{Setup.}
In this experiment, we test the hypothesis that most of the gradient should flow through the components that we judge \textit{a priori} to be most critical to the model's predictions.
We are particularly interested in whether the gradient flow through a Transformer matches our expectation of the self-attention mechanism's components.
So, while we compute the top gradient path from the output to the input representations, we only inspect the top path at a Transformer's main branching point, which is when the hidden state is passed into the skip connection and the keys, values, and queries of the self-attention mechanism (\cref{fig:bert_kvq}).
If we observe higher levels of gradients flowing through one branch, a natural interpretation is that this component is more critical for the model's prediction. 
To test whether this interpretation is justified, we construct a task where we can clearly reason about how a well-trained Transformer model ought to behave and identify how well the top gradient flow aligns with our expectations of a model's critical component.\looseness=-1

\paragraph{Model.} We use a 1-layer Transformer model with hidden layer size of \num{16} and \num{2} attention heads to minimize branching points and increase interpretability.
We train this model to achieve 100\% validation accuracy on the task described below.
\tikzset{%
    representation/.style={draw=none, fill=gray!40, text width=0.7cm, inner sep=3pt, minimum height=2cm, rounded corners},
    keysnode/.style={draw=none, fill=keycolor!80, text width=1.2cm, inner sep=3pt, rounded corners},
    valuesnode/.style={draw=none, fill=valuecolor, text width=1.2cm, inner sep=3pt, rounded corners},
    queriesnode/.style={draw=none, fill=querycolor, text width=1.2cm, inner sep=3pt, rounded corners},
    extranode/.style={draw=black, fill=gray!20, line width=0.3mm, rounded corners},
    bertgroup/.style={draw=black, rounded corners, dotted},
    label/.style={draw=none, text width=3cm, align=left, anchor=west},
    selfattngroup/.style={draw=black, fill=gray!10, rounded corners, dotted},
    background rectangle/.style={draw=white!15},
    show background rectangle,
    arrow/.style={line width=0.3mm}
}

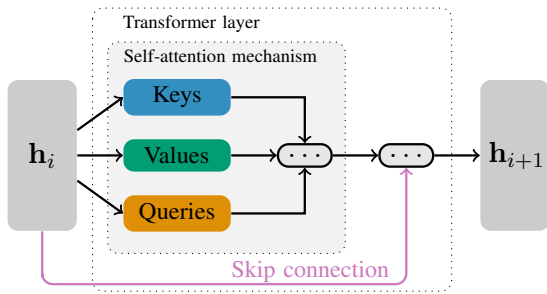
\begin{figure}
     \centering
    \begin{tikzpicture}[align=center,node distance=0.3cm]
        \node[representation] (input) {$\vh_i$};
        
        \node[valuesnode] (values) [right=0.6cm of input] {Values};
        \node[keysnode] (keys) [above=of values] {Keys};
        \node[queriesnode] (queries) [below=of values] {Queries};
    
        \node[extranode] (selfattn) [right=0.6cm of values] {$\ldots$};
        \node[extranode] (norm) [right=0.6cm of selfattn] {$\ldots$};
        
        \node[representation] (output) [right=0.6cm of norm] {$\vh_{i+1}$};

        \draw[bertgroup] ($(keys.north west)+(-0.4,0.85)$) rectangle ($(queries.south east)+(2.9,-0.7)$);
        \draw[selfattngroup] ($(keys.north west)+(-0.2,0.4)$) rectangle ($(queries.south east)+(1.5,-0.2)$);
        
        \node[label,font=\scriptsize] (selfattnlabel) [above=0.00cm of keys,xshift=0.8cm] {Self-attention mechanism};
        
        \node[label,font=\scriptsize] (bertgrouplabel) [above=0.40cm of keys,xshift=0.8cm] {Transformer layer};
        
        \node[label,font=\small,color=skipcolor] (skiplabel) [below=1.10cm of norm.south,xshift=-0.8cm] {Skip connection};

        \node[valuesnode,font=\small] (values) [right=0.6cm of input] {Values};
        \node[keysnode,font=\small] (keys) [above=of values] {Keys};
        \node[queriesnode,font=\small] (queries) [below=of values] {Queries};
        \node[extranode] (selfattn) [right=0.6cm of values] {$\ldots$};
    
        \draw[arrow,->] (input) -- (keys.west);
        \draw[arrow,->] (input) -- (values.west);
        \draw[arrow,->] (input) -- (queries.west);
        \draw[arrow,->, rounded corners, draw=skipcolor] (input.south) |- ++(0,-0.7cm) -| (norm.south); 
        \draw[arrow,->] (keys.east) -| (selfattn);
        \draw[arrow,->] (values.east) -- (selfattn);
        \draw[arrow,->] (queries.east) -| (selfattn);
        \draw[arrow,->] (selfattn.east) -- (norm.west);
        \draw[arrow,->] (norm.east) -- (output);
    \end{tikzpicture}
     
     \caption{Simplified computation graph for the $i^\text{th}$ layer of a Transformer~\citep{vaswani-transformer} encoder at the key branching point where a hidden state is passed into the \textcolor{skipcolor}{skip connection} and the \textcolor{keycolor}{keys}, \textcolor{valuecolor}{values}, and \textcolor{querycolor}{queries} of the self-attention mechanism.
     We use $\vh_i$ to denote the hidden representations returned by layer $i - 1$, and use ``$\ldots$'' to denote others parts of the computation graph.
     }
     \label{fig:bert_kvq}
     \vspace{-0.5cm}
 \end{figure}
 
\paragraph{Task.} 
We design the \texttt{FirstTokenRepeatedOnce} task to target the utility of this method for interpreting the self-attention mechanism. 
In this task, an input consists of a sequence of numbers, which is labeled according to whether the first token appears again at any point in the sequence, e.g., \texttt{[\colorbox{pink}{1}, 4, 6, \colorbox{pink}{1}] $\rightarrow$ True}, whereas \texttt{[\colorbox{pink}{3}, 4, 6, 2] $\rightarrow$ False}. 
Furthermore, the inputs are constrained such that the first token will be repeated at most once, to isolate the decision-making of the model to the presence (or lack thereof) of a single token. 
We randomly generate a dataset of \num{10000} points with sequence length \num{10} and vocab size \num{20}. 
The correct decision-making process for this task entails comparing the first token to all others in the sequence and returning \texttt{True} if there is a match.
This is, in fact, analogous to how queries and keys function within the self-attention mechanism: A query $q_t$ is compared to the key $k_{t'}$ of each token ${t'}$ in the sequence and the greater the match, the greater attention paid to token $t'$ by query token $t$.
We would therefore expect that the self-attention mechanism relies heavily on the query representation of the first token and key representations of the remaining tokens and, in particular, the key representation of the repeated token, if present.
In turn, we hypothesize the max-product gradient value will primarily originate from the queries branch for the first token and keys for the remaining tokens, and be especially high for the repeat token.

\paragraph{Results.}
The results, summarized in \Cref{fig:max_grad_synthetic}, provide strong evidence for our hypothesis that the behavior of the $(\max, \times)$ gradient reflects the importance of the different model components. 
We observe all expected gradient behaviors described in the previous paragraph, and especially that the highest gradient flow (for any token) is through the keys of the repeat token.

\subsection{Top gradient path of BERT for subject--verb agreement}
\label{sec:top_path_experiments}

\begin{figure*}[t]
     \centering
     \includegraphics[width=1.0\linewidth]{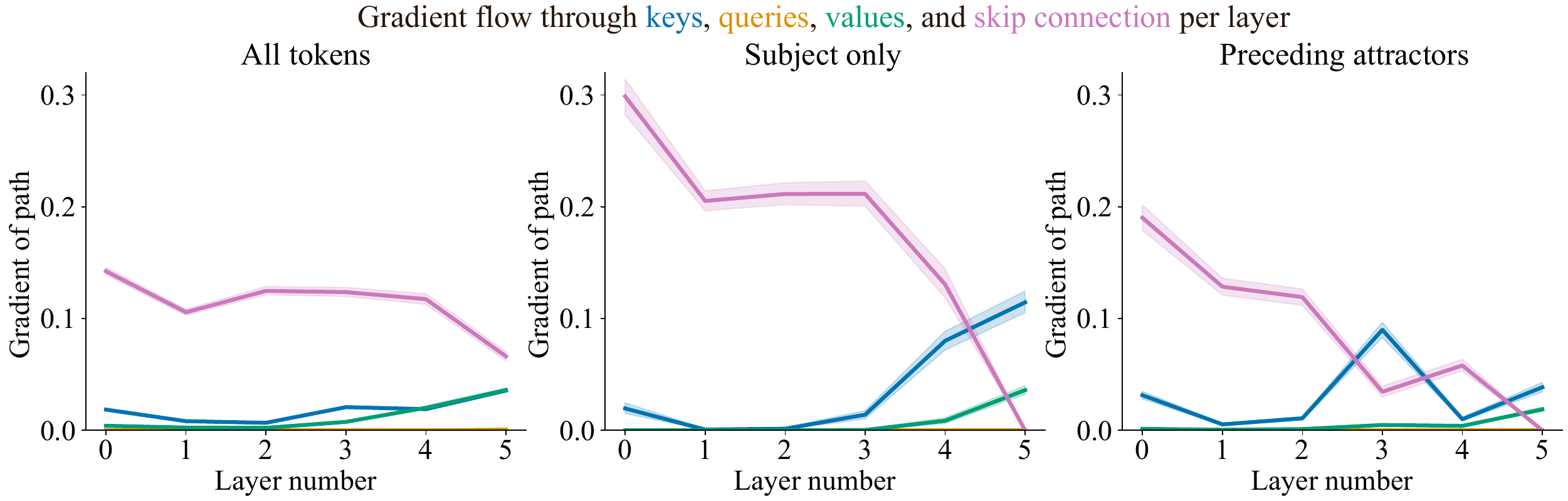} 
     \caption{Each plot depicts the top gradient path behavior of BERT for different tokens in a sentence, averaged across the \num{670} sentences. 
     Notably, for subjects (middle plot), the \textcolor{skipcolor}{\textit{skip connection}} contains the top gradient path for all layers except the final layer, which is consistent with findings from \citep{lu-etal-inf-patterns-2021}. 
     In the final layer, most of the gradient for both the subject and attractors flows through the \textcolor{keycolor}{\textit{keys}} of the self-attention mechanism. This differs from the gradient flow averaged across all tokens, indicating this behavior is specific to the nouns of the sentence.}
     \label{fig:max_grad_kvq}
     \vspace{-0.2cm}
 \end{figure*}

\paragraph{Setup.}
We now apply this method to understand the self-attention mechanism of a larger model (BERT) for the more complex NLP task of SVA.
We subsample \num{1000} examples from the dataset from \citet{linzen_assessing_2016} and use spaCy \citep{matthew_spacy_2020} to identify the subject and attractors within each sentence. We then filter down to \num{670} sentences after removing sentences where BERT tokenizes the subject or attractors as multiple tokens.
Using the max-product semiring, we then compute the top gradient path through the different branches (skip connection, keys, values, and queries) for (a) the subject of a sentence, (b) the attractors of a sentence, and (c) all tokens of a sentence.

\paragraph{Model.}
BERT~\citep{bert} is a popular encoder-only Transformer model for many NLP tasks.
BERT's architecture consists of multiple Transformer encoder layers stacked atop each other, along with a task-specific head. 
We use the \texttt{google/bert\_uncased\_L-6\_H-512\_A-8} pretrained model from Huggingface~\citep{wolf_transformers_2020}, which has \num{6} attention layers, hidden size of \num{512}, and \num{8} attention heads.

\paragraph{Task.} 

We consider the subject--verb number agreement task in our experiments.
Variants of this task in English have become popular case studies in neural network probing. 
Notably, this phenomenon has been used to evaluate the ability for models to learn hierarchical syntactic phenomena \cite{linzen_assessing_2016,gulordava-etal-2018-colorless}. It has also served as a testing ground for interpretability studies which have found evidence of individual hidden units that track number and nested dependencies \cite{lakretz-etal-2019-emergence}, and that removing individual hidden units or subspaces from the models' representation space have a targeted impact on model predictions \cite{finlayson-etal-2021-causal,lasri-etal-2022-probing}.
Our formulation of the task uses BERT's native masked language modeling capability by recasting it as a cloze task: We mask a verb in the sentence and compare the probabilities with which BERT predicts the verb forms with correct and incorrect number marking. For example, given the input ``all the other albums produced by this band \mask their own article,'' we compare the probabilities of ``have'' (correct) and ``has'' (incorrect).
We compute the gradient with respect to the difference between the log probability of the two inflections.\looseness=-1

The data for this experiment is from \citet{linzen_assessing_2016}. All the examples in their dataset also include one or more \defn{attractors}. These are nouns such as ``band'' in the example above, which (a) are not the subject, (b) precede the verb, and (c) disagree with the subject in number. Furthermore, all masked verbs are third person and present tense, to ensure that number agreement is non-trivial.

\paragraph{Results.}
From \cref{fig:max_grad_kvq}, we highlight key differences between the $(\max, \times)$ gradient behavior for subject tokens and all tokens in general. 
Most saliently, for subject tokens only, the max-product gradient flows entirely through the self-attention mechanism in the last layer and mostly through the skip connection in earlier layers, which is consistent with findings from \citet{lu-etal-inf-patterns-2021}.
Moreover, within the self-attention mechanism, most (76\%) of the gradient in the last layer for the subject flows through the keys matrix.
In contrast, across all tokens, the top gradient paths mostly through the skip connection for all layers, and otherwise is more evenly distributed between keys and values.

We also note similarities and differences between the gradient flows of the subject and preceding attractors.
Both exhibit a similar trend in which the gradient flows primarily through the keys (and entirely through the self-attention mechanism) in the last layer. 
However, the top gradient has a greater magnitude for the subject than the attractors (especially in the keys). 
Since self-attention uses a token's keys to compute the relative importance of that token to the \mask token, we speculate that the max-product gradient concentrating primarily on the keys (and more so for the subject than attractors) reflects that a successful attention mechanism relies on properly weighting the importances of the subject and attractors.

\subsection{Gradient graph entropy vs.\ task difficulty}
\label{sec:entropy_experiments}

\begin{figure}[t]
     \centering
     \includegraphics[width=1.0\linewidth]{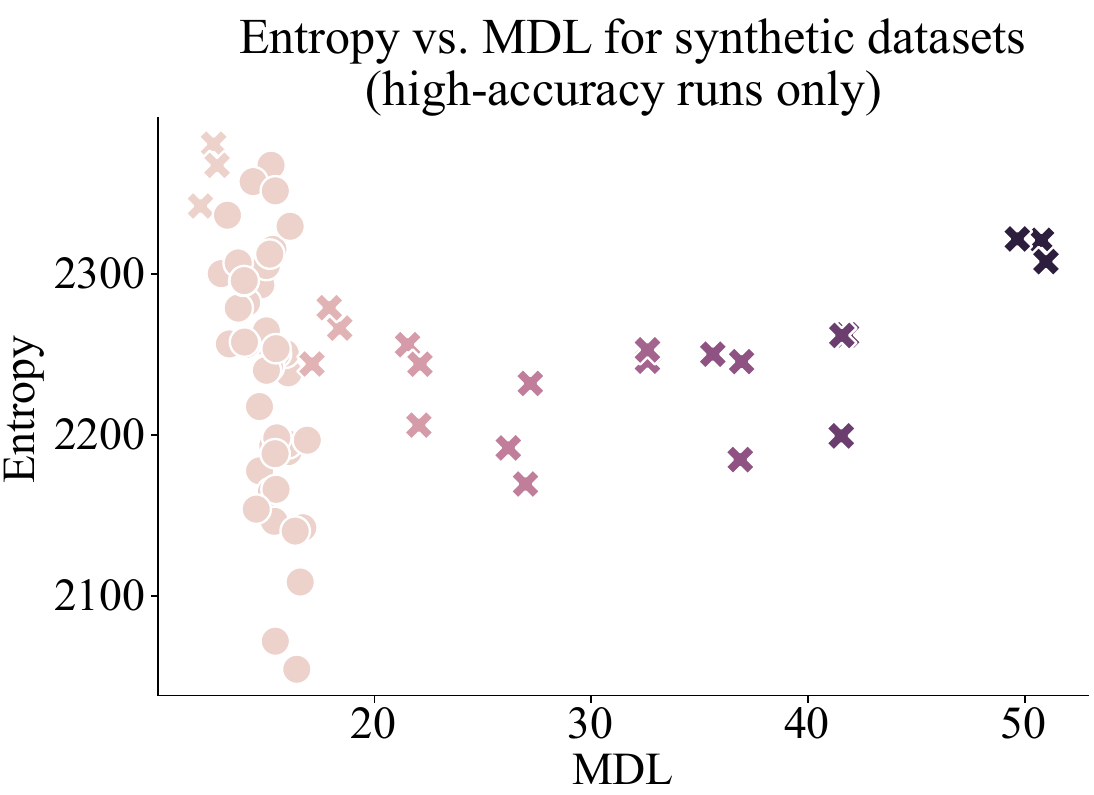} 
     \caption{Each point represents the entropy and MDL of a model trained on a given dataset (3 seeds per dataset).
     We denote the \texttt{BinCountOnes} datasets with the $\thicctimes$ marker and other tasks (\texttt{ContainsTokenSet} and tasks from \citealp{lovering2021predictinginductive}) with the $\bullet$ marker.
     The hue corresponds to the number of classes of the task; the lightest hue indicates a binary problem while the darker hues indicate more classes (max of 36).
     }
     \label{fig:entropy_mdl_combined}
     \vspace{-0.5cm}
 \end{figure}

\paragraph{Setup.} This experiment tests the hypothesis that the entropy of a model's gradient graph is positively correlated with the difficulty of the task that the model was trained to solve.
We construct a variety of synthetic tasks and compare the average gradient entropy of a 2-layer transformer on examples in each of these tasks.
We measure the difficulty of a task with the minimum description length 
\citep[MDL;][]{rissanen1978modeling}.\footnote{The MDL of a dataset under a model measures the number of bits required to communicate the labels of the dataset, assuming the sender and receiver share both the unlabeled data and a model, which can be used to reduce the information the sender must transmit. 
Alternatively, MDL can be thought of as the area under the loss curve as a function of dataset size.}
Following the approach used by \citet{lovering2021predictinginductive} and \citet{voita-titov-2020-information}, we measure MDL by repeatedly training the model on the task with increasing quantities of data and summing the loss from each segment.
The higher the MDL, the more difficulty the model had in extracting the labels from the dataset, and therefore the more challenging the task.
We hypothesize that a model will have higher entropy for more difficult tasks because it will require using more paths in its computation graph.
During our analysis, we drop runs where the model was unable to achieve a validation accuracy of $>90\%$, to avoid confounding results with models unable to learn the task.

\paragraph{Model.} For all tasks, we use the same \num{2}-layer transformer architecture with a hidden layer size of \num{64}, \num{4} attention heads, and always predicts a distribution over \num{36} classes (with some possibly unused); this ensures our results are comparable across tasks with different numbers of classes.
We train the models for \num{50} epochs on each of the synthetic datasets.\looseness=-1

\paragraph{Task.} We design a variety of synthetic tasks in order to control for difficulty more directly.
In the \texttt{ContainsTokenSet} family of tasks, an input is a sequence of $S$ numbers and labeled \texttt{True} or \texttt{False} based on whether the input contains \textit{all} tokens in a pre-specified token set. 
Different tasks within \texttt{ContainsTokenSet} are defined by the pre-specified token set.   
The \texttt{BinCountOnes} family of tasks is parameterized by a number of classes $C$. In this task, an input $x$ is a sequence of $S$ numbers. 
The label $y$ is determined by the number of 1s in the sequence according to the following function:
$y(x) = \ceil*{\frac{\texttt{Count1}(x)}{S/C}} - 1$,
i.e., in the 2-class instance of \texttt{BinCountOnes}, an input is labeled 0 if it contains $\leq S/2$ 1s and 1 if it contains $>S/2$ 1s.
Finally, we also evaluate on the synthetic datasets 
\texttt{Contains1}, \texttt{AdjacentDuplicate},
\texttt{FirstTokenRepeatedImmediately}, and \texttt{FirstTokenRepeatedLast} from \citep{lovering2021predictinginductive}.
For more details, see \cref{app:datasets}.

\paragraph{Results.} 
The results show clear evidence against our initial hypothesis that gradient entropy increases as a function of task difficulty, as measured by MDL. 
While there appears to be some patterns evident between entropy and MDL in \Cref{fig:entropy_mdl_combined}, their interpretation is unclear.
From observing the lightest-hued points there appears to be a negative linear relationship between entropy and MDL for the binary tasks.
However, confusingly, the $\thicctimes$ points seem to suggest a quadratic-like relationship between entropy and MDL for the \texttt{BinCountOnes} tasks.
We speculate that this could be explained by a phase-change phenomena in the model's learning dynamics. 
That is, for sufficiently easy tasks, the model need not focalize much in order to solve the task. 
Incrementally more difficult tasks may require the model to focalize more, thus resulting in the decreasing entropy for tasks below a certain MDL threshold.
Then, once a task is sufficiently difficult, the model is required to use more of the network to solve the task. 
Therefore, we see this increase in entropy as the MDL increases past a certain threshold for the \texttt{BinCountOnes} task.
The presence of these clear (although somewhat mystifying) patterns indicates that there exists \textit{some} relationship between entropy and MDL. 
More experimentation is needed to understand the relationship between entropy and MDL for task difficulty.

\section{Conclusion}

We presented a semiring generalization of the backpropagation algorithm, which allows us to obtain an alternative view into the inner workings of a neural network.
We then introduced two semirings, the max-product and entropy semirings, which provide information about the branching points of a neural network and the dispersion of the gradient graph.
We find that gradient flow reflects model component importance, gradients flowing through the self-attention mechanism for the subject token pass primarily through the keys matrix, and the entropy has some relationship with the difficulty of learning a task.
Future work will consider semirings outside the scope of this work,
e.g., the \defn{top-$k$ semiring} \citep{goodman1999semirings} to track the top-$k$ gradient paths, as well as computing semirings online for control during training. \looseness=-1

\section{Limitations}

While our approach inherits the linear runtime complexity of the backpropagation algorithm, runtime concerns should not be fully neglected. 
Firstly, the linear runtime is only an analytical result, not an empirical measure. This means that the actual runtime of the backpropagation and thus our algorithm depend heavily on their implementation. 
For instance, some deep learning frameworks do a better job at reusing and parallelizing computations than others \cite{goodfellow2016deep}. 
Indeed, our code is optimized for good readability and extensibility at the expense of speed, which hints at another limitation of our approach:
Our approach requires deep integration with the framework as it needs access to all model weights and the computation graph. 
For this reason, our approach cannot be easily packaged and wrapped around any existing model or framework and we instead developed our own JAX-based reverse-mode autodifferentiation library, based on the numpy-based Brunoflow library \citep{Brunoflow}.
While we release our library to enable other researchers to analyze models through their gradient graphs, it faces some computational and memory constraints.
In our experiments, running the three semirings together on a single sentence can take several minutes (depending on sentence length) using \texttt{google/bert\_uncased\_L-6\_H-512\_A-8}, the 6-layered pretrained BERT from Huggingface \citep{wolf_transformers_2020}, totaling our experimentation time on our datasets at about 10 CPU-hours.
For improved adoption of this method, we encourage the direct integration of semiring implementations into the most popular deep learning frameworks. 
Our final point pertains not only to our study but to most interpretability approaches: One has to be careful when drawing conclusions from gradient paths. 
Cognitive biases, wrong expectations, and omitted confounds may lead to misinterpretation of results.

\section*{Ethics statement}
We foresee no ethical concerns with this work. Our work aims to make the inner workings of neural network models more interpretable. On this account, we hope to contribute to reducing biases inherent in model architectures, pre-trained model weights, and tasks by increasing overall transparency. 

\section*{Acknowledgements}

Kevin Du acknowledges funding from the Fulbright/Swiss Government Excellence Scholarship.
Lucas Torroba Hennigen acknowledges support from the Michael Athans fellowship fund. 
Niklas Stoehr acknowledges funding through the Swiss Data Science Center (SDSC) Fellowship. Alex Warstadt acknowledges support through the ETH Postdoctoral Fellowship program.

\bibliography{references/anthology, references/custom, references/nik_refs}
\bibliographystyle{acl_natbib}

\appendix
\newpage
\onecolumn

\section{Implementation of Top Gradient Path}\label{app:top_grad_path}

In practice, we implement the top gradient path by storing 4 additional fields to each node in the graph: the most positive gradient of the node, a pointer to the child node which contributed this most positive gradient, the most negative gradient of the node, and a pointer to the child node which contributed this most negative gradient. In this way, each node tracks the paths containing the most positive gradient ($\toppos$) and most negative gradient ($\topneg$) from itself to the output node. To dynamically extend the path from $v_k$ to $v_j$ ($j < k$):
\begin{align*}
      v_j.\toppos &= \begin{cases}
        v_k.\toppos \cdot \deriv{v_k}{v_j} & \text{if } \deriv{v_k}{v_j} \geq 0
        \\
        v_k.\topneg \cdot \deriv{v_k}{v_j} & \text{otherwise}
        \end{cases} \\
      v_j.\topneg &= \begin{cases}
        v_k.\topneg \cdot \deriv{v_k}{v_j} & \text{if } \deriv{v_k}{v_j} \geq 0
        \\
        v_k.\toppos \cdot \deriv{v_k}{v_j} & \text{otherwise}
        \end{cases}
\end{align*}

\section{Additional Entropy Sanity Checks and Experiments}
\subsection{Sanity Checks with Synthetic Data}
\begin{figure}[h]
\centering
\begin{subfigure}{.5\textwidth}
  \centering
  \includegraphics[width=1.0\linewidth]{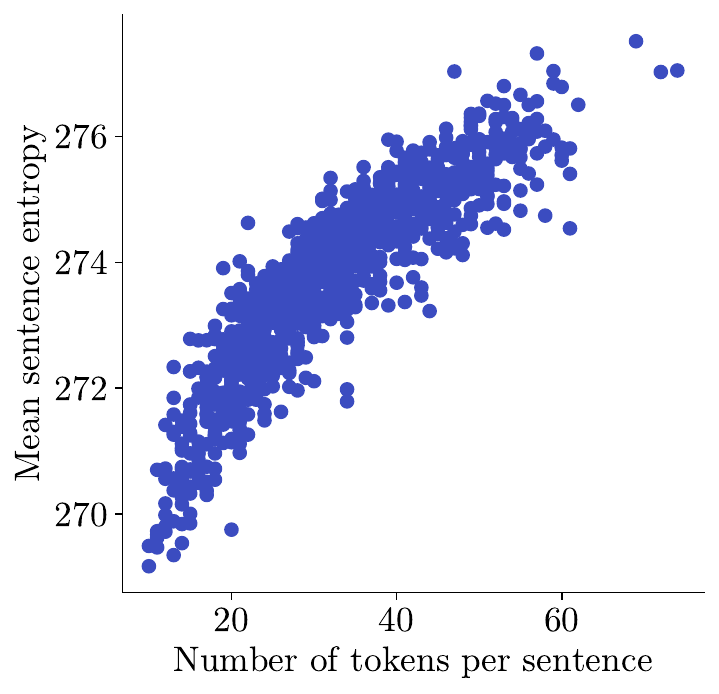}
  \caption{Entropy vs sentence length}
  \label{fig:entropy_sanity_sentence_length}
\end{subfigure}%
\begin{subfigure}{.5\textwidth}
  \centering
  \includegraphics[width=1.0\linewidth]{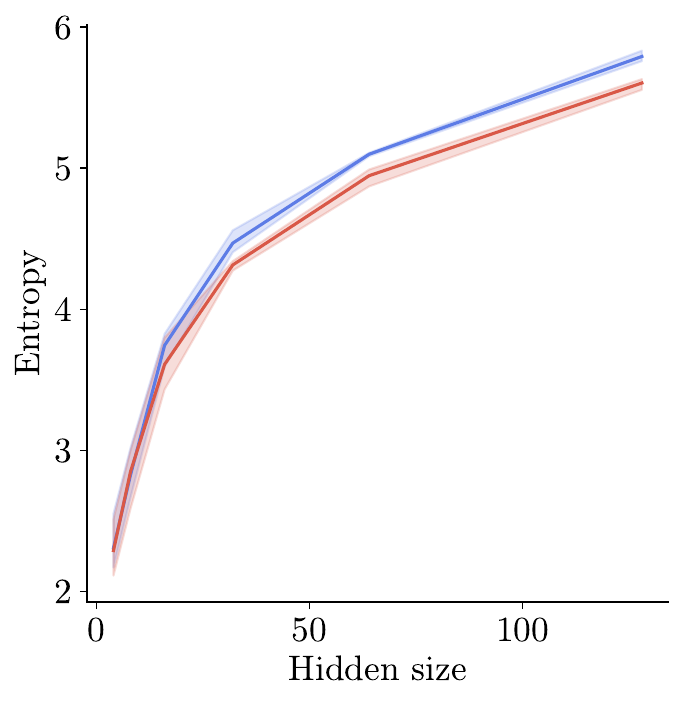}
  \caption{Entropy vs model complexity (\textcolor{blue_fig}{relevant} \& \textcolor{red_fig}{irrelevant} features)}
  \label{fig:entropy_sanity_model_complexity}
\end{subfigure}
\caption{\Cref{fig:entropy_sanity_sentence_length} shows that the more tokens a sentence contains, the more gradient paths are naturally involved and consequently the higher the overall entropy. 
\Cref{fig:entropy_sanity_model_complexity} shows that as the hidden size of the MLP increases, so too does the model entropy. By comparing the entropy of the  \textcolor{blue_fig}{relevant feature line} and the \textcolor{red_fig}{irrelevant features line}, it also appears that the entropy is consistently higher for the relevant feature than irrelevant features, especially as model complexity increases.}
\label{fig:entropy_sanity}
\end{figure}

To build intuition about the entropy of a model's computation graph, we run two sanity check experiments. First, we evaluate the entropy of a pretrained BERT model as the sentence length increases. 
Since larger sentence lengths result in more paths in the computation graph, we expect the entropy of the model to increase with sentence length. 
Our findings confirm this (\Cref{fig:entropy_sanity_sentence_length}).

Second, we expect that the entropy of a trained model ought to increase with the model complexity, as measured by hidden size.
In this experiment, we create a 4-featured artificial dataset with randomly generated values in the range $[0, 1]$, labeled by whether the first feature is greater than 0.5. 
We train multilayer perceptrons with varying hidden sizes on this dataset and find that the entropy of the input features increases with model complexity as expected (see \cref{fig:entropy_sanity_model_complexity}).

\subsection{Entropy vs Example Difficulty in Subject--Verb Agreement}
\paragraph{Setup.} We investigate the relationship between the entropy of the gradient graph of BERT and input sentences in the task of subject--verb number agreement. 
In this task, we measure example difficulty by the number of attractors in a sentence (more attractors corresponds to greater difficulty). We sub-sample the dataset from \citet{linzen_assessing_2016} to \num{1000} sentences, balanced evenly by the number of attractors per sentence (ranging from 1 to 4 attractors). 
Then, using the entropy semiring, we compute the entropy of BERT's gradient graph for each sentence. 

\begin{figure}[h]
     \centering
     \includegraphics[width=0.6\linewidth]{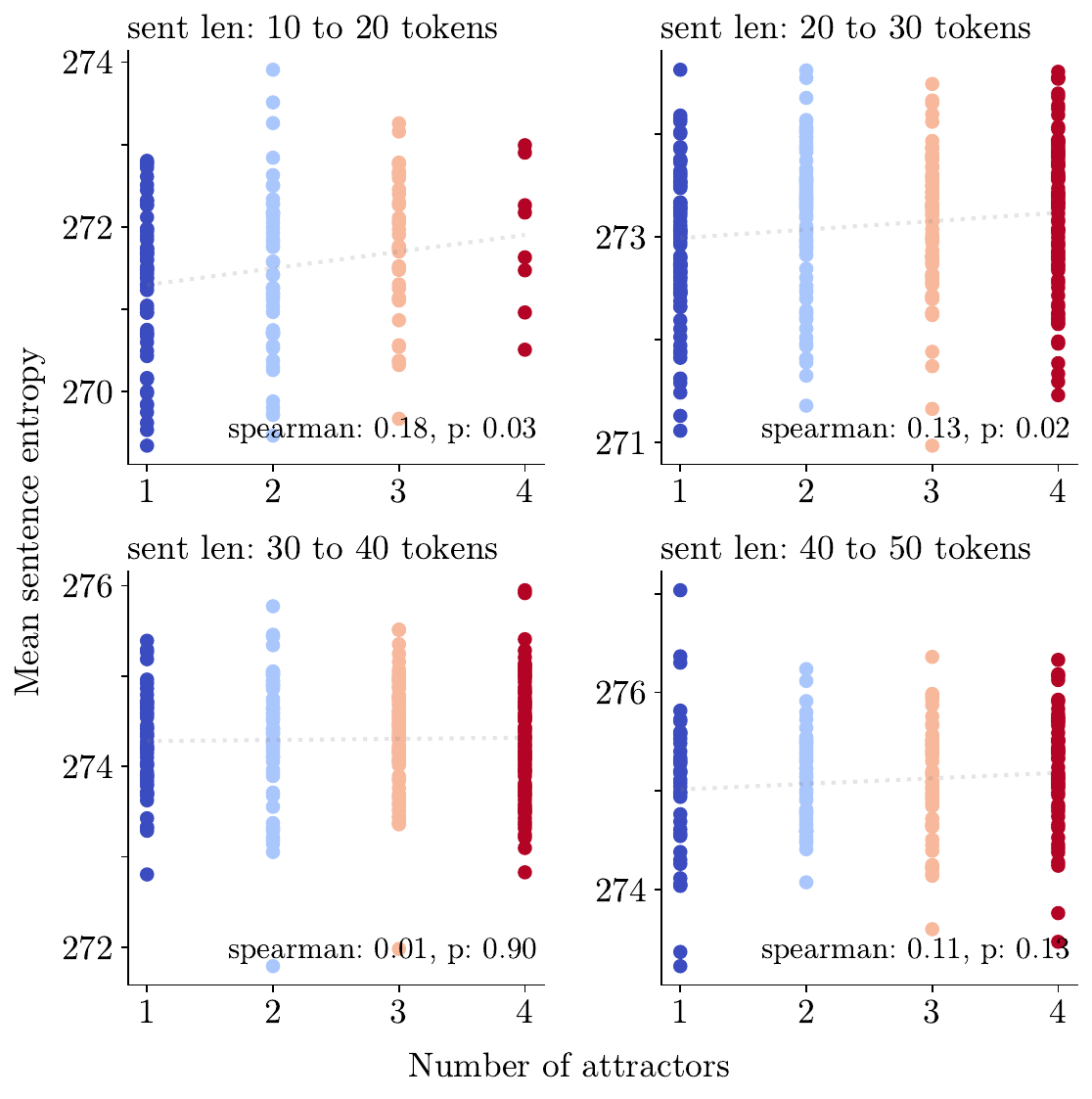} 
     \caption{For all sentence length bins, there appears to be little to no correlation between number of attractors and entropy.}
     \label{fig:entropy_attractors}
\end{figure}

\paragraph{Results.}
Since sentences with more tokens will naturally have a higher entropy due to a larger computation graph (see \cref{fig:entropy_sanity_sentence_length}), we control by sentence length. 
We bin sentences of similar length for (10--20, 20--30, 30--40, and 40--50 tokens) before analyzing the effect that the number of attractors has on entropy. 
We present the results in \cref{fig:entropy_attractors} and additionally run a Spearman correlation test between the entropy of the input representations (averaged across all tokens in the sentence) and the number of attractors. 
For each group of sentence lengths, we find minimal correlation between number of attractors and entropy. 
Therefore, there is little evidence to support a relationship between entropy and example difficulty as measured by number of attractors.
However, number of attractors is not necessarily a strong indicator of example difficulty, and recommend more rigorous comparison of entropy against a stronger metric of example difficulty in future work.

\section{Synthetic Datasets}
\label{app:datasets}
\subsection{Binary Datasets}
We list in \Cref{tab:binary_datasets} descriptions and examples of all binary tasks constructed for our experiments.
\label{app:binary_datasets}
\begin{table}[h]
\renewcommand{\arraystretch}{1.5}
\begin{tabular}{@{}p{3cm} p{2.0cm} p{4.0cm} p{2.7cm} p{2.8cm}@{}}
\toprule
Task Name                     & Parameterized by:                     & Description                                                                                                                                                                                                                               & Positive Example  & Negative Example  \\ \midrule
ContainsTokenSet              & A set of tokens, $T$, e.g., \texttt{\{1,2,3\}} & Labeled \texttt{True} if $X$ contains every token in $T$ and \texttt{False} otherwise                                                                                                                   & \texttt{[1,3,4,2,5,2]} & \texttt{[1,5,9,2,2,4]} \\
Contains1                     & N/A                                   & Labeled \texttt{True} if $X$ contains the token \texttt{1} and \texttt{False} otherwise                                                                                                & \texttt{[1,3,4,2,5,2]} & \texttt{[6,5,9,2,2,4]} \\
FirstToken-RepeatedImmediately  & N/A                                   & Labeled \texttt{True} if the first two tokens in $X$ are the same and \texttt{False} otherwise                                                                                                          & \texttt{[3,3,2,6,7,8]} & \texttt{[5,3,2,6,7,8]} \\
FirstToken-RepeatedLast        & N/A                                   & Labeled \texttt{True} if the first and last tokens in $X$ are the same and \texttt{False} otherwise                                                                                                     & \texttt{[8,3,2,6,7,8]} & \texttt{[8,3,2,6,7,4]} \\
AdjacentDuplicate             & N/A                                   & Labeled \texttt{True} if two adjacent tokens in $X$ are the same at any point in the sequence and \texttt{False} otherwise                                                                              & \texttt{[1,3,6,6,7,8]} & \texttt{[1,3,6,8,7,8]} \\
FirstToken-RepeatedOnce        & N/A                                   & Labeled \texttt{True} if the first token in $X$ is repeated at any point in the sequence and \texttt{False} otherwise. $X$ is further constrained to have at most one repeat of the first token in $X$. & \texttt{[1,3,6,1,7,8]} & \texttt{[1,3,6,7,7,8]} \\ \bottomrule
\end{tabular}
\caption{Binary synthetic datasets used in \Cref{sec:top_path_synthetic_experiments} and \Cref{sec:entropy_experiments}.
For all tasks, the input $X$ is a sequence of $S$ numbers (valued from 1 to vocab size).
While for the examples in this table we use $S=6$ to save space, in the actual experiments we use $S=10$ (\Cref{sec:top_path_synthetic_experiments}) and $S=36$ (\Cref{sec:entropy_experiments}).}
\label{tab:binary_datasets}
\end{table}

\subsection{BinCountOnes Datasets}
\label{app:multiclass_datasets}
We construct one family of multiclass classification datasets, \texttt{BinCountOnes}.

\paragraph{Parameterization.} A \texttt{BinCountOnes} task is parameterized by the number of classes $C$, between 2 to $S$, such that $C$ divides $S$. For example, when $S=6$, $C$ could be 3.

\paragraph{Description.} Each example $X$ is labeled between $[0, C-1]$ by the following formula: $\mathrm{label}(X) = \ceil*{\frac{\mathrm{Count1}(X)}{S/C}} - 1$, where $C\mathrm{Count1}(X)$ is the number of 1s that appear in $X$.

\paragraph{Examples.} See \Cref{tab:bincountones_examples}.
\begin{table}[h!]
\begin{tabular}{@{}ll@{}}
\toprule
Input         & Label \\ \midrule
\texttt{[1,3,4,2,5,2]} & 0     \\
\texttt{[1,3,4,2,3,1]} & 0     \\
\texttt{[1,3,4,2,1,1]} & 1     \\
\texttt{[1,3,1,1,5,1]} & 1     \\
\texttt{[1,3,1,1,1,1]} & 2     \\
\texttt{[1,1,1,1,1,1]} & 2     \\ 
\bottomrule
\end{tabular}
\caption{Example inputs and labels for the \texttt{BinCountOnes} task where sequence length $S=6$ and number of classes $C=3$.}
\label{tab:bincountones_examples}
\end{table}

\end{document}